\documentclass[runningheads]{llncs}
\usepackage[T1]{fontenc}
\usepackage{graphicx}
\graphicspath{{figures/}}
\usepackage{booktabs}   %
\usepackage{multirow}   %
\usepackage{amsmath}    %
\usepackage{amssymb}    %
\usepackage[table]{xcolor}  %
\usepackage[hidelinks]{hyperref}
\usepackage{orcidlink}
\renewcommand\orcidID[1]{\,\orcidlink{#1}}
\usepackage{cleveref}
\usepackage{url}  %
\usepackage{placeins}  %
\newcommand\megadl{Mega\-De\-scrip\-tor-L}

\newcommand\our{{WildMatch}}

\begin{document}
\title{\texorpdfstring{\raisebox{-0.3em}{\includegraphics[height=1.4em]{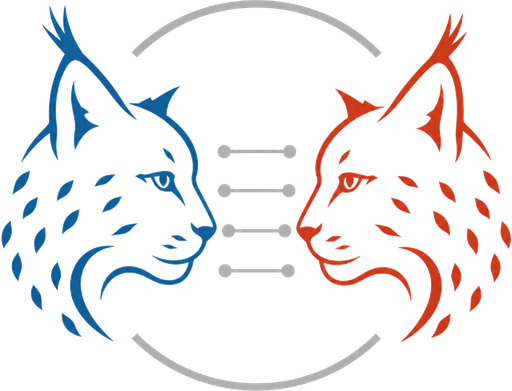}}~}{}WildMatch: Weakly Supervised Image Matcher Adaptation for Wildlife
Re-Identification}
\titlerunning{WildMatch: Weakly Supervised Image Matcher Adaptation}
\newif\ifanon
\anonfalse  %
\ifanon
\author{Anonymous Author(s)}
\authorrunning{Anonymous Author(s)}
\institute{Anonymous Institution(s)}
\else
\author{Turhan Can Kargin\inst{1,2}\orcidID{0000-0002-6751-4773}\thanks{T.\,C. Kargin and P. Kubaty contributed equally.} \and
Piotr Kubaty\inst{1,2} \and
Ekaterina Rostovskaya\inst{2,5}\orcidID{0000-0002-8421-1838} \and
Izabela Wierzbowska\inst{5}\orcidID{0000-0001-6329-7241} \and
Bartosz Zieli\'nski\inst{1,3}\orcidID{0000-0002-3063-3621} \and
Marcin Przewi\k{e}\.{z}likowski\inst{1,4}\orcidID{0000-0003-4772-3268}}
\authorrunning{T.\,C. Kargin, P. Kubaty et al.}
\institute{Faculty of Mathematics and Computer Science,
Jagiellonian University, Poland \and
Doctoral School of Exact and Natural Sciences,
Jagiellonian University, Poland \and
Jagiellonian Center for Artificial Intelligence, Krak\'ow, Poland \and
NASK National Research Institute, Warsaw, Poland \and
Institute of Environmental Sciences, Faculty of Biology, Jagiellonian University, Krak\'ow, Poland %
\\
\email{\{turhancan.kargin, piotr.kubaty, e.rostovskaya\}@doctoral.uj.edu.pl}\\
\email{\{bartosz.zielinski, marcin.przewiezlikowski, i.wierzbowska\}@uj.edu.pl}}
\fi
\maketitle              %
\begin{abstract}

Individual animal re-identification from camera-trap imagery is an instance retrieval problem central to non-invasive wildlife monitoring: a query image must retrieve the correct individual from a reference set of known animals.
This requires computer vision models to recognize distinctive local patterns in fur, skin, or other visual markings. 
Current approaches either learn global embeddings as a classification problem, requiring many labeled images per individual while largely ignoring local evidence, or apply off-the-shelf, domain-agnostic image matchers. 
Although such matchers are pretrained on large and diverse image collections, adapting them to wildlife imagery is challenging because available datasets are small and lack correspondence-level annotations.
We study weakly supervised adaptation of a pretrained keypoint matcher using only identity labels, without keypoint-level or geometric correspondence ground truth. 
We mine informative image pairs with the pretrained matcher, derive weak positive and negative supervision from identity agreement, and contrastively fine-tune the matching network to strengthen correspondences for same-identity pairs and suppress them for different identities.
Across open-source wildlife re-identification datasets, our approach improves accuracy over off-the-shelf matchers and a state-of-the-art local--global fusion method. %
Under an open-world protocol with held-out individuals, it learns a transferable correspondence prior rather than memorizing training identities.
To our knowledge, this is the first study of matcher-level, identity-supervised adaptation for animal re-identification. 
Our method enables data-efficient specialization of image matching models to wildlife domains using identity annotations already available in typical monitoring datasets.
Our project page can be found at \href{https://wildmatch.gmum.net}{\texttt{wildmatch.gmum.net}}.

\end{abstract}

\section{Introduction}

\begin{figure}[t]
\centering
\includegraphics[width=\textwidth]{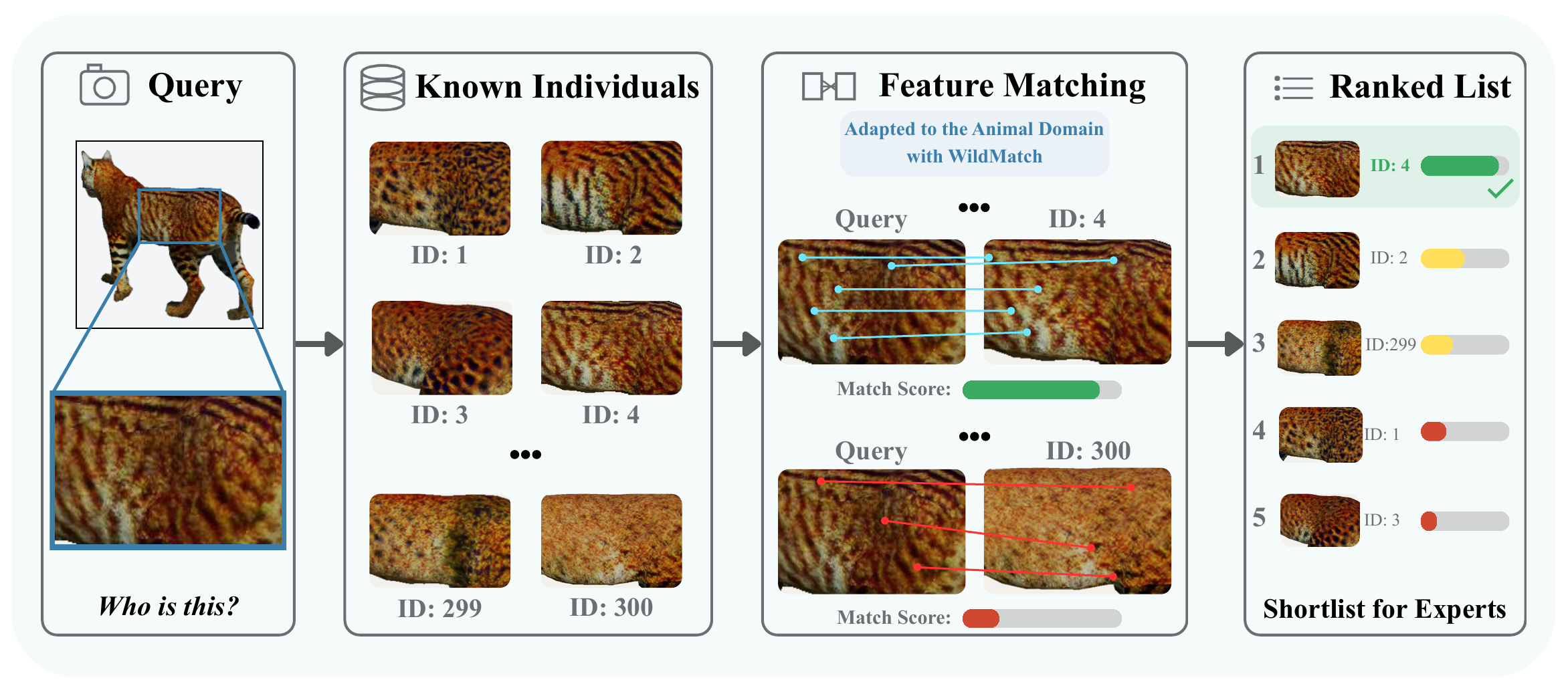}
\caption{\textbf{Wildlife re-identification as instance retrieval.} A
camera-trap query is compared with every known individual, and a local
feature matcher ranks them by the summed confidence of point pairs that
both images agree on, giving a short list for an expert. The database
holds hundreds of individuals, of which six are shown. \our{} performs
weakly supervised adaptation of the matcher using only the identity
labels of the database (Fig.~\ref{fig:pipeline}), without keypoint or
correspondence annotation. Images are synthetic CzechLynx
renders~\cite{picek2025czechlynx}, with illustrative matches and scores.}
\label{fig:teaser}
\end{figure}

Ecological studies on wildlife monitoring and conservation require knowledge of demographic parameters, such as population size and structure.
Whereas non-invasive monitoring methods, such as camera-trapping, allow for minimized anthropogenic disturbance, estimating animal densities and turnover rates depends on capture-recapture modelling~\cite{dula}.
Individuals with explicit and unique coat patterns get distinguished, and their long-term well-being can be assessed~\cite{czarnota}.
Recognizing an individual means deciding which known animal a new photograph shows. Individual animal re-identification can therefore be formulated as an instance retrieval problem, where a query image must match the correct individual from a reference set of known animals, as shown in~\Cref{fig:teaser}.
When recognizing individuals, human experts rely on distinctive local markings, such as the arrangement of spots on a lynx's coat~\cite{czarnota}.
They use extensive opportunistic camera-trapping or dual camera-trap setup to assign both body flanks to one individual and to compare coat patterns between multiple individuals~\cite{dula,suessle}.

This naturally motivates local feature matching as a computational approach to wildlife re-identification~\cite{wildfusion2025}.
Global image embeddings discard much of this local evidence, since in visual models spatial information resides mainly in patch-level features and is largely obscured by global pooling~\cite{kargin2026sparrta}.
A pairwise matcher has no identity-specific parameters and can in principle learn visual evidence that remains useful for previously unseen individuals~\cite{nordstrom2026lomalocalfeaturematching}.
Adapting such a matcher to a particular species has the potential to further improve identification accuracy, but presents practical challenges.
Wildlife re-identification datasets usually provide image-level identity labels, whereas local matchers are conventionally trained using geometric or keypoint-level correspondence supervision~\cite{Chen_2025_CVPR}.
An identity label establishes that two images depict the same individual, but does not reveal which local structures correspond between them.
Moreover, due to bilateral asymmetry of flanks, two images of the same animal may contain little or no shared local evidence~\cite{suessle}.
The central question we address is therefore: \textbf{Can a general-purpose image matcher be specialized to a wildlife domain using only the sparse identity annotations already available in re-identification datasets?}

We introduce \textbf{\our{}}, a weakly supervised approach for adapting pretrained local image matchers using identity labels alone.
Our key idea is to use the pretrained matcher itself to identify image pairs for which informative local evidence is likely to exist.
For each training image, we mine high-scoring images of the same individual as positive pairs and high-scoring images of different individuals as hard negatives.
Identity agreement then provides weak pair-level supervision without requiring manual annotation of keypoints or geometric correspondences.
We contrastively fine-tune the matching module, encouraging higher correspondence scores for same-identity pairs and lower scores for different-identity pairs.
Intuitively, the pretrained matcher identifies where correspondence is plausible, while identity labels determine whether such correspondence should be encouraged or suppressed.

We evaluate \our{} across eight wildlife re-identification datasets spanning multiple species and substantially different visual domains.
Our experiments show that identity-supervised adaptation improves over both off-the-shelf local matchers and a state-of-the-art local--global fusion approach across most datasets and candidate budgets (see~\Cref{fig:main-top5}).
We further investigate where the limited identity supervision is most effectively used, finding that adapting the matching module improves both evaluated matcher architectures, whereas applying the same supervision to the descriptor branch does not provide the same benefit.
Finally, an open-world evaluation on individuals completely absent from matcher training shows that the gains persist beyond the identities used for adaptation, indicating that \our{} learns a transferable correspondence prior rather than a closed-set identity classifier.

\textbf{Our contributions can be summarized as follows:}
\begin{itemize}
    \item We introduce \our{}, a weakly supervised method for adapting pretrained local image matchers to wildlife re-identification using only identity labels, without keypoint-level or geometric correspondence annotations.
    \item We demonstrate across eight wildlife re-identification datasets that adapting the matching module improves identification accuracy over off-the-shelf matchers and competitive local--global baselines.
    \item We show that the learned adaptation transfers to previously unseen individuals and analyze where identity supervision should be applied within the matching pipeline.
\end{itemize}

\begin{figure}[htb]
    \centering
    \includegraphics[width=0.95\textwidth]{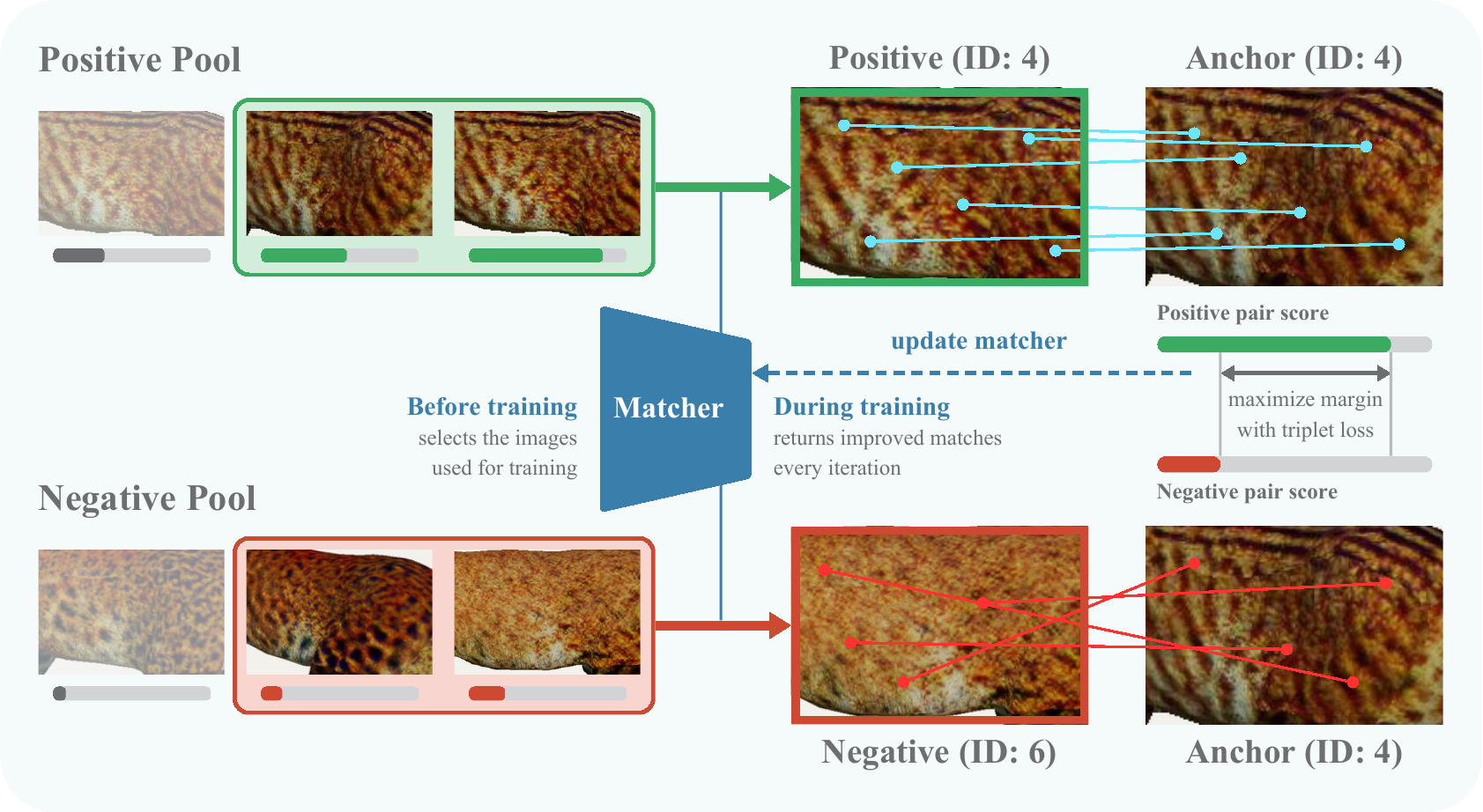}
    \caption{\textbf{Weakly supervised matcher adaptation with \our{}.}
        Before training, the pretrained matcher selects top-scoring positives and hard
        negatives. A triplet loss then fine-tunes the matcher to score positive pairs
        above hard negatives.
    }
    \label{fig:pipeline}
\end{figure}

\section{Related Work}
\label{sec:related}

\textbf{Local feature matching} establishes correspondences between salient image regions and underlies tasks such as retrieval, localization, and 3D reconstruction~\cite{lowe2004sift}.
Modern learned pipelines combine local feature extraction with neural matching modules that reason jointly over an image pair.
LightGlue~\cite{lindenberger2023lightgluelocalfeaturematching} introduced an efficient attention-based matcher, RDD~\cite{Chen_2025_CVPR} improves robustness of the underlying detector and descriptor, and LoMa~\cite{nordstrom2026lomalocalfeaturematching} shows further gains from scaling data, model capacity, and training.
These methods target general-purpose geometric matching, whereas we specialize matching to visual cues distinguishing individual animals.

\textbf{Photographic animal re-identification} has long supported non-invasive mark--recapture studies, with early computer-assisted systems exploiting distinctive visual markings~\cite{bolger2012markrecapture}.
Recent approaches predominantly learn global image representations, ranging from species-specific methods such as EDEN~\cite{chelak2021edendeepfeaturedistribution} to multi-species models~\cite{cermak2024wildlifedatasets,otarashvili2024multispeciesanimalreidusing}.
Standardized resources such as WildlifeDatasets~\cite{cermak2024wildlifedatasets}, WildlifeReID-10k~\cite{adam2025wildlifereid10k}, and CzechLynx~\cite{picek2025czechlynx} expose substantial variation in pose, illumination, image quality, and observations per individual.

\textbf{Local matching for animal re-identification} complements global representations by explicitly comparing distinctive markings.
HotSpotter~\cite{crall2013hotspotter} uses keypoint matching for patterned species, while related work also addresses viewpoint-specific issues such as identifying the visible flank~\cite{suessle}.
WildFusion~\cite{wildfusion2025} combines global similarity with pretrained local matchers, and recent AnimalCLEF systems likewise employ LightGlue-based local verification~\cite{smith2026dsgtarcanimalclef2026}.
In these approaches, however, the matcher itself remains general-purpose rather than being adapted using animal identity labels.

\textbf{Feature matcher adaptation} has recently shown that pretrained matching modules can be further specialized.
Wang~\cite{wang2026understandingoptimizingattentionbasedsparse} fine-tunes attention-based sparse matchers across different local feature detectors while retaining pretrained feature representations.
This work targets compatibility across feature extractors, whereas \our{} adapts the matching module to wildlife imagery using only image-level identity agreement, without geometric, pose, or keypoint-level correspondence supervision.

\section{Method}
\label{sec:method}

Current wildlife re-identification systems often employ image matchers to compare a query with candidate reference images retrieved from an annotated database.
These matchers are typically neural-network-based and pretrained on general imagery~\cite{wildfusion2025,otarashvili2024multispeciesanimalreidusing}.

\our{} adapts such pretrained local feature matchers to the wildlife domain using only image-level identity labels.
Starting from a wildlife re-identification dataset labeled just with animal identities, we use the pretrained matcher to mine informative positive and hard-negative image pairs.
We then fine-tune its matching module with a triplet ranking objective~\cite{hoffer2018deepmetriclearningusing}, as illustrated in \Cref{fig:pipeline}.
Our method applies to pretrained matchers with a learnable matching module that can be adapted while retaining the original local feature representation.

\subsection{Problem Setup and Inference}
\label{sec:pipeline}
\label{sec:matcher}
\label{sec:data-prep}

Let $\mathcal{Q}$ denote the query set and $\mathcal{G}$ the gallery, with identity label $y(x)$ associated with each image $x$.
For a query $q\in\mathcal{Q}$, the goal is to rank gallery images depicting the same individual above those depicting other individuals.

For each image $x$, the local feature extraction stage identifies $n_x$ keypoints with coordinates $\mathbf{p}_i^x\in\mathbb{R}^2$ and produces their local descriptor vectors $\mathbf{z}_i^x\in\mathbb{R}^d$.
The matching module $M_\theta$ takes the coordinates and descriptors from two images $(x,x')$ and predicts a confidence matrix $\mathbf{P}_\theta(x,x')\in[0,1]^{n_x\times n_{x'}}$.
Its entry $P_{ij}$ scores the candidate correspondence between keypoint $i$ in $x$ and keypoint $j$ in $x'$.
The matcher retains a keypoint pair as a correspondence if the two keypoints select each other as their highest-confidence match and $P_{ij}$ exceeds a confidence threshold.
These retained pairs form the correspondence set $\mathcal{A}_\theta(x,x')$.
We aggregate their confidences into an image-level matching score:
\begin{equation}
s_\theta(x,x')=
\frac{1}{\min(n_x,n_{x'})}
\sum_{(i,j)\in\mathcal{A}_\theta(x,x')}
P_{ij}.
\label{eq:score}
\end{equation}
The score combines the number and confidence of retained correspondences, normalized by the smaller keypoint count.
At inference, we evaluate $s_\theta(q,g)$ for each image $g$ in a candidate set $\mathcal{C}_k(q)\subseteq\mathcal{G}$.
Candidates are re-ranked by decreasing matching score, and the highest-ranked results are returned.

In practice, we construct $\mathcal{C}_k(q)$ as the $k$ gallery images with the highest cosine similarity between embeddings produced by a global image encoder, such as DINOv3~\cite{siméoni2025dinov3} or MegaDescriptor~\cite{cermak2024wildlifedatasets}.
To reduce uninformative non-animal keypoints and bias arising from stationary camera traps repeatedly capturing the same backgrounds, we additionally mask background pixels before feature extraction using dataset-provided animal segmentation masks; when these are unavailable, the animal is segmented with SAM~3~\cite{carion2026sam3segmentconcepts} with dataset-specific prompts ordered from species names to broader
animal-oriented descriptions.

\subsection{Identity-Guided Pair Mining}
\label{sec:pair-mining}

Training \our{} requires positive pairs that share informative local evidence and challenging negative pairs.
Identity labels alone do not identify such pairs: images of the same individual may show different viewpoints or opposite sides of the body and contain little or no shared visible markings.
We therefore use the pretrained matcher to select high-scoring same-identity pairs as positives and high-scoring different-identity pairs as hard negatives.

Let $\mathcal{D}_{\mathrm{train}}$ denote the labeled training set and $\theta_0$ the pretrained matcher parameters.
We perform pair mining once, before fine-tuning, using the matching score $s_{\theta_0}$ from Eq.~\eqref{eq:score}.
Pair mining is performed directly on $\mathcal{D}_{\mathrm{train}}$, independently of the global retrieval stage used at inference.
To reduce redundancy and imbalance between frequent and rarely observed individuals, we subsample the training images and organize them into identity-specific collections.
Each retained image is treated as an anchor $a$ and matched against the remaining eligible training images.

For each anchor, we form pools $\mathcal{P}(a)$ and
$\mathcal{N}(a)$ containing images of the same individual
and other individuals, respectively.
Using $s_{\theta_0}(a,\cdot)$, we select up to five high-scoring
images for each pool, distributing the selected examples across
eligible collections where multiple collections are available.
This favors positives with plausible shared local evidence and
negatives that the pretrained matcher already finds visually similar.
Anchors without an eligible positive are discarded.
Finally, we retain a limited number of anchors per collection, prioritizing those with the highest-scoring pair in either pool,
$
\max_{u\in\mathcal{P}(a)\cup\mathcal{N}(a)}
s_{\theta_0}(a,u),$
which limits the influence of disproportionately large collections.

\subsection{\our{}: Identity-Supervised Matcher Adaptation}
\label{sec:finetune}

We train the matcher to assign higher scores to 
image pairs of the same identity than to pairs of different identities.
For each anchor $a$, we sample $p$ uniformly from $\mathcal{P}(a)$
and $n$ from a mixture of hard negatives in $\mathcal{N}(a)$
and random images of other training individuals to increase
sample diversity.
Starting from $\theta_0$, we update only the matching module, while keeping the local feature extractor and global encoder frozen.

The inference score in Eq.~\eqref{eq:score} relies on discrete correspondence selection and confidence thresholding, which may discard all correspondences for a training pair.
We therefore optimize a relaxed score computed directly from $\mathbf{P}_\theta(x,x')$ before these selection steps.
For each keypoint, we take the confidence of its highest-scoring candidate in the other image and average these values symmetrically across both images:
\begin{equation}
\tilde{s}_\theta(x,x')=
\frac{1}{2}\left(
\frac{1}{n_x}\sum_{i=1}^{n_x}\max_j P_{ij}
+
\frac{1}{n_{x'}}\sum_{j=1}^{n_{x'}}\max_i P_{ij}
\right).
\label{eq:relaxed}
\end{equation}
This provides a training signal even when no correspondence survives the inference-time selection filters.

We optimize the matcher using a triplet margin loss:
\begin{equation}
\mathcal{L}(\theta)=
\mathbb{E}_{(a,p,n)}\left[
\bigl[
\alpha
-\tilde{s}_\theta(a,p)
+\tilde{s}_\theta(a,n)
\bigr]_+
\right],
\label{eq:loss}
\end{equation}
where $[u]_+=\max(0,u)$ and $\alpha=0.5$.
The objective encourages the adapted matcher to assign stronger correspondence evidence to same-identity pairs than to different-identity pairs.
\section{Experimental Setup}
\label{sec:setup}

\begin{figure}[b]  %
\centering
\includegraphics[width=\textwidth]{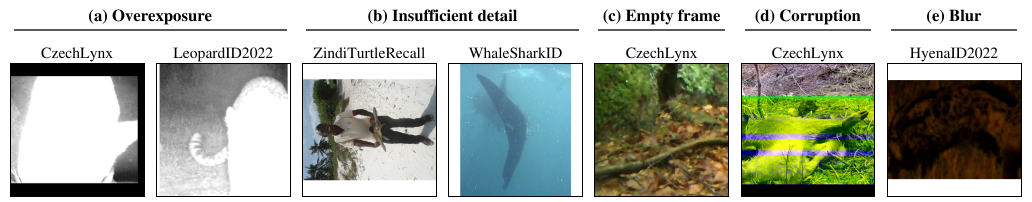}
\caption{\textbf{Examples of challenging images in the benchmark datasets.} Raw
frames that are overexposed (a), show too little of the animal to see its
markings (b), contain no visible animal (c), are corrupted (d), or are
blurred (e). Such images remain in the database and query sets of all
methods.}
\label{fig:data-quality}
\end{figure}

\paragraph{Datasets.}\label{sec:datasets} We evaluate \our{} on eight public datasets covering diverse species
(Table~\ref{tab:datasets}). Six of them come from the WildlifeReID-10k
collection~\cite{adam2025wildlifereid10k,cermak2024wildlifedatasets}, the
seventh is CzechLynx~\cite{picek2025czechlynx}, a recent camera-trap dataset
of Eurasian lynx, and the eighth is SalamanderID2025, a collection of fire
salamander photographs from the AnimalCLEF 2025
challenge~\cite{adam2025overview}. We use the official splits, except for
Salamander, whose query labels are not public, so we hold out the latest
capture date of each individual photographed on at least two dates as
queries. The database split is used for fine-tuning, and query images are
never seen during training. As is typical of field data, many images are
noisy (Fig.~\ref{fig:data-quality}), and we do not filter them. The
datasets are also highly imbalanced (Table~\ref{tab:datasets}), which
motivates the balanced metric below.

\begin{table}[htb]
\centering
\caption{\textbf{Dataset statistics.} Database (DB) images form the retrieval gallery and are the only images used for fine-tuning, and query (Q) images are evaluated against them. Med.\ and Max give the median and maximum number of database images per individual. The small plot shows the image counts of all individuals, sorted from most to least photographed, with the top 10\% in dark blue, and Top 10\% gives their share of all database images. Single is the share of individuals with one image.}
\label{tab:datasets}
\begin{tabular}{lrrrrrrcrr}
\toprule
 & \multicolumn{2}{c}{Images} & \multicolumn{2}{c}{Individuals} & \multicolumn{5}{c}{Images per individual} \\
\cmidrule(lr){2-3}\cmidrule(lr){4-5}\cmidrule(lr){6-10}
Dataset & DB & Q & DB & Q & Med. & Max & Distribution & Top 10\% & Single \\
\midrule
Nyala~\cite{dlamini2020automated} & 1{,}514 & 428 & 237 & 236 & 5 & 25 & \raisebox{-0.5ex}{\includegraphics[height=2.6ex,width=6em]{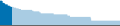}} & 30\% & 24\% \\
Hyena~\cite{lilawphyena_2022} & 2{,}499 & 630 & 256 & 256 & 6 & 72 & \raisebox{-0.5ex}{\includegraphics[height=2.6ex,width=6em]{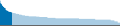}} & 35\% & 2\% \\
Leopard~\cite{lilawpleopard_2022} & 5{,}373 & 1{,}433 & 430 & 380 & 3 & 329 & \raisebox{-0.5ex}{\includegraphics[height=2.6ex,width=6em]{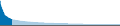}} & 70\% & 31\% \\
Sea star~\cite{wahltinez_wahltinez_2024} & 1{,}759 & 428 & 95 & 95 & 14 & 30 & \raisebox{-0.5ex}{\includegraphics[height=2.6ex,width=6em]{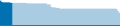}} & 15\% & 0\% \\
Whale shark~\cite{holmberg_norman_arzoumanian_2009} & 6{,}108 & 1{,}585 & 543 & 512 & 6 & 123 & \raisebox{-0.5ex}{\includegraphics[height=2.6ex,width=6em]{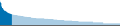}} & 42\% & 14\% \\
Turtle~\cite{zindi_2022} & 9{,}721 & 3{,}082 & 2{,}265 & 2{,}251 & 2 & 92 & \raisebox{-0.5ex}{\includegraphics[height=2.6ex,width=6em]{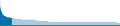}} & 39\% & 3\% \\
\midrule
Salamander~\cite{adam2025overview} & 1{,}138 & 246 & 584 & 221 & 1 & 11 & \raisebox{-0.5ex}{\includegraphics[height=2.6ex,width=6em]{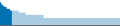}} & 31\% & 64\% \\
\midrule
CzechLynx~\cite{picek2025czechlynx} & 27{,}836 & 11{,}924 & 319 & 319 & 26 & 2{,}465 & \raisebox{-0.5ex}{\includegraphics[height=2.6ex,width=6em]{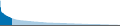}} & 61\% & 0\% \\
\bottomrule
\end{tabular}
\end{table}

\paragraph{Protocols.}\label{sec:protocols} All methods use the candidate
budgets $k \in \{10, 50, 100, 250, 500, 1000\}$, with $k{=}250$ as the
default. The WildlifeReID-10k datasets and the closed split of CzechLynx
follow the closed-set protocol, in which every query individual is
present in the database. To test transfer to individuals never seen
during training, we also define an \emph{unseen-identity protocol} on
CzechLynx. It uses the 44 individuals that appear in the test part of the
open split but not in its training part. For each of them, the earliest
encounter forms the gallery and all later encounters form the queries,
which yields 160 gallery and 2{,}081 query images and rules out
encounter-level leakage. The matcher is fine-tuned on the training part
of the open split (275 individuals), so none of the 44 contributes a
training pair. Since
the gallery holds only 160 images, the budgets are
$k \in \{10, 50, 100, 160\}$, and $k{=}160$ is exhaustive matching.

\paragraph{Baselines and Metrics.}\label{sec:baselines} We compare against four groups of baselines, all evaluated on the same
background-masked images and splits. \emph{Cosine retrieval} ranks the
database by the cosine similarity of global embeddings, either from
\megadl{}~\cite{cermak2024wildlifedatasets}, an encoder pretrained
on animal re-identification data, or from the general-purpose
DINOv3~\cite{siméoni2025dinov3}. \emph{Off-the-shelf matchers} rank
the $k$ candidates with the unadapted LoMa or RDD checkpoints and isolate
the effect of adaptation. \emph{WildFusion}~\cite{wildfusion2025} fuses
calibrated global and local similarity scores over the same $k$
candidates. A \emph{linear probe} trains an identity classifier with a
class-weighted loss on \megadl{} with the backbone frozen, partially,
or fully fine-tuned, testing
if the identity labels are better spent on the global model. Since a domain expert typically inspects a short list of candidates, our
primary metric is \emph{top-5 accuracy}, the fraction of queries 
correctly matched to one of the
five highest-ranked identities. We also report
\emph{balanced top-1 accuracy}, the top-1 accuracy averaged over rare and frequent
individuals, weighted equally.

\paragraph{Implementation Details.}\label{sec:impl} We fine-tune only the
matching module (11.9M parameters) and keep the detector and descriptor
frozen. Inputs are 512-pixel images with up to 512 keypoints. Triplets are
mined once on the training images with the pretrained matcher
(Sec.~\ref{sec:pair-mining}). We train for 300 epochs with AdamW, a batch
size of 32, a learning rate of $10^{-5}$ with a cosine schedule, a weight
decay of $10^{-4}$, and gradient clipping at 1, and report the final
checkpoint. On CzechLynx, fine-tuning takes 5.1 GPU-hours on an NVIDIA
RTX~4090 (24\,GB).

\section{Experimental Results}
\label{sec:results}

\subsection{Retrieval Accuracy}
\label{sec:main-results}
\begin{figure}[t]
\centering
\includegraphics[width=0.95\textwidth]{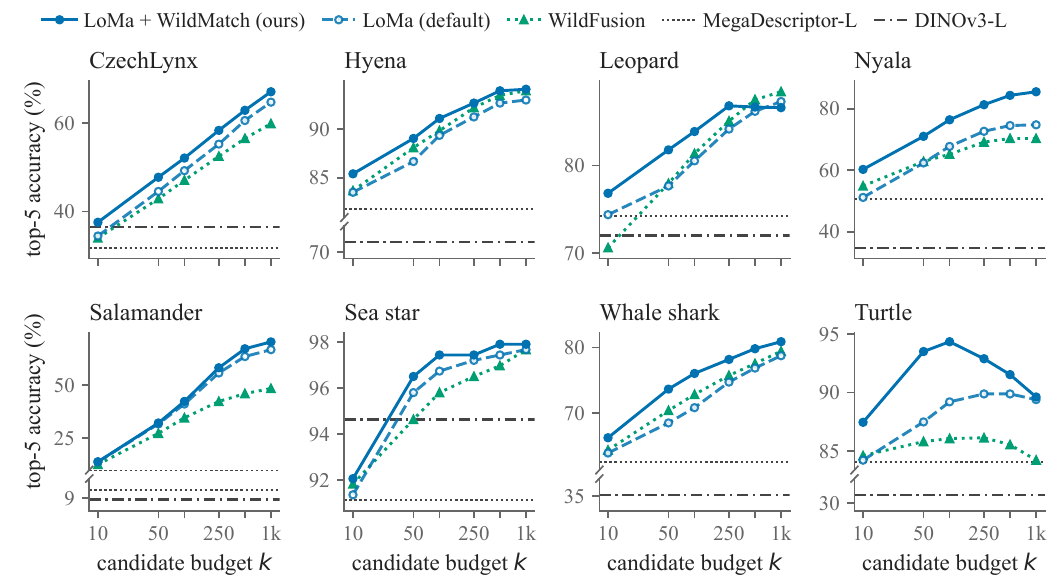}
\caption{\textbf{Retrieval accuracy on eight datasets.} Top-5 accuracy as a function of the candidate budget $k$ (log
scale). LoMa + WildMatch (ours) is compared with the default LoMa
matcher, WildFusion, and cosine retrieval with \megadl{}, which
also provides the candidates for all matching methods, and with DINOv3-L
(flat lines).}
\label{fig:main-top5}
\end{figure}

We first test whether the adapted matcher retrieves the correct individual
more often than the default matcher and WildFusion, and how this depends
on the number of candidates $k$ that the matcher scores. Figure~\ref{fig:main-top5} reports top-5 accuracy as a function of the
candidate budget $k$. Across the eight datasets, the fine-tuned matcher
improves over its default checkpoint and over WildFusion at almost every
budget. Fine-tuning also
saves compute. To reach the accuracy of the fine-tuned matcher at a given
$k$, the default matcher needs a candidate list two to five times longer.
Re-ranking with any matcher lifts accuracy well above embedding retrieval
alone. \megadl{} is the stronger first stage on most datasets,
while DINOv3-L is better on CzechLynx and Sea star.

\subsection{Which Component Should Be Adapted?}
\label{sec:where-to-adapt}
A matcher consists of a detector and descriptor network, which produces
keypoints and their features, and a matching module, which assigns
correspondences between them. Table~\ref{tab:descriptor} fine-tunes each
part, and both together, on the closed CzechLynx split with the same
objective and training pairs. Fine-tuning the descriptor branch lowers the
balanced top-1 accuracy, whereas
fine-tuning the matching module improves both matchers. Identity labels
indicate which images should match, which is what the matching module
scores, while the descriptors were trained for geometric repeatability and
may lose it under an identity objective. Fine-tuning both parts is slightly
better than the matching module alone, but 16 to 21 times more expensive to train, since a trainable descriptor rules out
caching the features and changes the input of the matching module at every
step. We therefore adapt only the matching module in all other experiments.

\begin{table}[t]  %
\centering
\caption{\textbf{Descriptor or matching module.} For each matcher, the descriptor branch, the matching module, or both are fine-tuned with the same identity supervision and training data (CzechLynx, closed split, $k{=}250$). Bal.\ is balanced top-1 accuracy, $\Delta$ its change from the default checkpoint, and GPU-h the training cost in GPU-hours of training steps. The matching-module row (ours) is shaded.}
\label{tab:descriptor}
\setlength{\aboverulesep}{0pt}\setlength{\belowrulesep}{0pt}\renewcommand{\arraystretch}{1.25}
\begin{tabular}{lrrrr<{\hspace{0.5em}}|>{\hspace{0.5em}}rrrr}
\toprule
 & \multicolumn{4}{c|}{LoMa} & \multicolumn{4}{c}{RDD-LightGlue} \\
\cmidrule(lr){2-5}\cmidrule(lr){6-9}
Fine-tuned part & Top-5 & Bal. & $\Delta$ & GPU-h & Top-5 & Bal. & $\Delta$ & GPU-h \\
\midrule
none (default) & 55.2 & 31.3 & -- & -- & 56.0 & 33.4 & -- & -- \\
descriptor branch & 56.1 & 28.6 & $-$2.6 & 77 & 49.8 & 25.7 & $-$7.7 & 89 \\
\cellcolor{green!12}matching module & \cellcolor{green!12}58.3 & \cellcolor{green!12}34.7 & \cellcolor{green!12}+3.4 & \cellcolor{green!12}5.1 & \cellcolor{green!12}56.8 & \cellcolor{green!12}34.4 & \cellcolor{green!12}+0.9 & \cellcolor{green!12}5.0 \\
both & 59.9 & 34.9 & +3.6 & 84 & 59.0 & 35.7 & +2.2 & 103 \\
\bottomrule
\end{tabular}
\end{table}

\subsection{Matcher Adaptation or Identity Classification}
\label{sec:where-to-spend}
\begin{figure}[t]
\centering
\includegraphics[width=\textwidth]{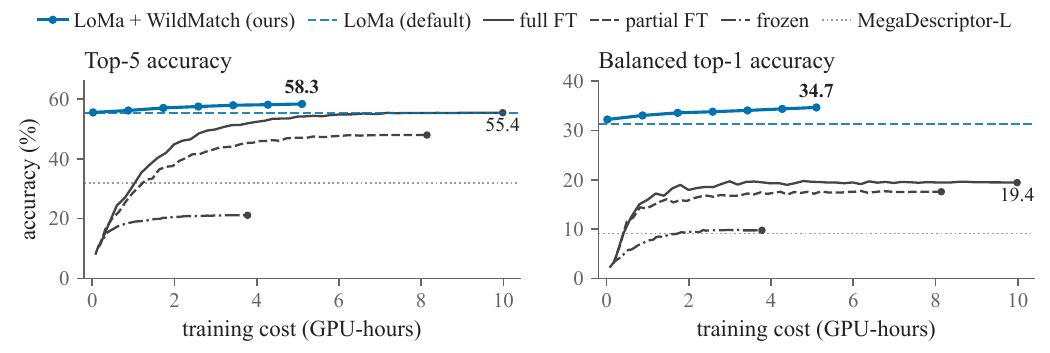}
\caption{\textbf{\our{} and classification by training
cost} (CzechLynx, closed split, $k{=}250$). Top-5 (left) and balanced top-1
accuracy (right) of LoMa + \our{} checkpoints and of the
\megadl{} classifier with the backbone frozen, partially fine-tuned
(partial FT), or fully fine-tuned (full FT), as a function of GPU-hours on
the same hardware.
The default LoMa matcher and cosine retrieval need no
training and are drawn as flat lines.}
\label{fig:training-cost}
\end{figure}

The same identity labels could instead be spent on the global model.
Figure~\ref{fig:training-cost} compares our adapted matching module with
the identity classifiers of Sec.~\ref{sec:baselines} as a function of
training cost on CzechLynx. After 10 GPU-hours of full fine-tuning, the
classifier only reaches the top-5 accuracy of the unadapted matcher and
stays below 20\% balanced top-1 accuracy, whereas the matcher starts at
31\% before adaptation and reaches 35\% after 5 GPU-hours. A classifier learns parameters for each
individual, so its training is dominated by the frequently photographed
ones, while the matcher is trained on image pairs and has no per-identity
parameters. The classifier also cannot recognize individuals outside its
label set, so it must be retrained whenever a new individual is added and
cannot be evaluated under the unseen-identity protocol
(Sec.~\ref{sec:unseen}).

\newpage  %
\subsection{Generalization to Unseen Identities}
\label{sec:unseen}
We evaluate both matchers under the unseen-identity protocol (Sec.~\ref{sec:protocols}).
Table~\ref{tab:unseen} shows that both fine-tuned matchers improve over
their default checkpoints and over WildFusion at every budget and in
both metrics. The gain persists at exhaustive $k{=}160$, where the first
stage plays no role, so the adaptation learns a transferable,
species-specific matching prior rather than the identities it was
trained on. An identity classifier cannot be applied here, since these
individuals have no class.

\begin{table}[b]  %
\centering
\caption{\textbf{Unseen-identity protocol on CzechLynx} (Sec.~\ref{sec:protocols}). Top-5 and balanced top-1 accuracy (\%) at candidate budgets $k$. Cosine retrieval does not depend on $k$. Fine-tuned matchers (ours) are shaded; best value per column in bold.}
\label{tab:unseen}
\begin{tabular}{lrrrrrrrr}
\toprule
 & \multicolumn{2}{c}{$k{=}10$} & \multicolumn{2}{c}{$k{=}50$} & \multicolumn{2}{c}{$k{=}100$} & \multicolumn{2}{c}{$k{=}160$} \\
\cmidrule(lr){2-3}\cmidrule(lr){4-5}\cmidrule(lr){6-7}\cmidrule(lr){8-9}
Method & Top-5 & Bal. & Top-5 & Bal. & Top-5 & Bal. & Top-5 & Bal. \\
\midrule
MegaDescriptor-L cosine & 19.6 & 12.0 & 19.6 & 12.0 & 19.6 & 12.0 & 19.6 & 12.0 \\
DINOv3-L cosine & 20.2 & 11.1 & 20.2 & 11.1 & 20.2 & 11.1 & 20.2 & 11.1 \\
\midrule
WildFusion & 21.7 & 17.3 & 27.7 & 21.1 & 30.6 & 22.5 & 32.6 & 23.1 \\
LoMa (default) & 22.7 & 18.9 & 30.9 & 23.0 & 35.5 & 25.8 & 40.4 & 26.7 \\
\rowcolor{green!12}
LoMa + \our{} & \textbf{25.0} & \textbf{19.7} & \textbf{36.6} & \textbf{27.3} & \textbf{40.9} & \textbf{30.3} & \textbf{46.1} & \textbf{31.8} \\
RDD (default) & 20.9 & 17.4 & 28.4 & 22.2 & 31.9 & 25.0 & 35.2 & 26.7 \\
\rowcolor{green!12}
RDD + \our{} & \underline{22.9} & \underline{19.6} & \underline{34.4} & \underline{25.9} & \underline{38.3} & \underline{28.7} & \underline{41.2} & \underline{29.7} \\
\bottomrule
\end{tabular}
\end{table}

\subsection{Applicability Beyond LoMa}
\label{sec:rdd-limits}
\begin{figure}[t]  %
\centering
\includegraphics[width=\textwidth]{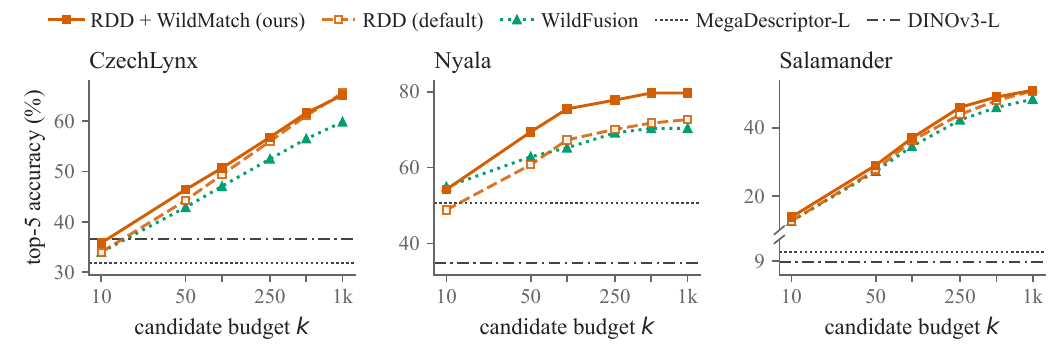}
\caption{\textbf{Generality to a second matcher.} Top-5 accuracy of RDD + WildMatch (ours) and the
default RDD-LightGlue matcher as a function of the candidate budget $k$
(log scale) on three datasets, with WildFusion and cosine retrieval for
reference. The advantage of fine-tuning is largest on Nyala and smaller on
Salamander and CzechLynx.}
\label{fig:rdd-k}
\end{figure}

The recipe is not specific to LoMa and applies to other local feature
matchers. We repeat the experiment with RDD and its LightGlue matcher,
which differ from LoMa in detector, descriptor, and training data.
Figure~\ref{fig:rdd-k} shows that fine-tuning the matching module again
improves over the default checkpoint. At the largest budgets the
advantage narrows. We attribute this to the training pairs. Most
negatives are hard negatives, the other-identity images that the
pretrained matcher scores highest, so the adapted matcher sees few of the
easier but more numerous candidates that a long list adds.

\subsection{Qualitative Results}
\label{sec:qualitative}
\begin{figure}[htb]  %
\centering
\includegraphics[width=\textwidth]{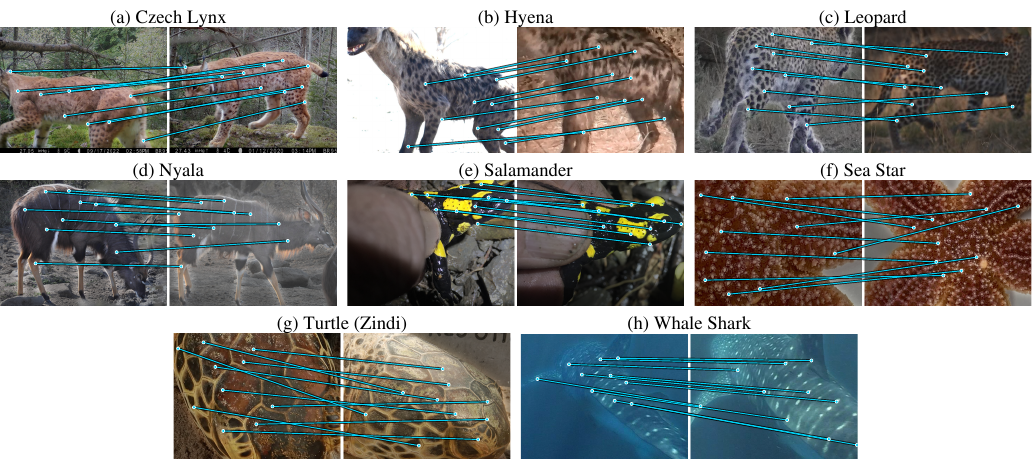}
\caption{\textbf{Qualitative results of top-1 retrievals with LoMa + \our{}}. For each dataset, a query (left) and its top-1 gallery image
(right) of the same individual. Of the 350 to 420 matches found for each
pair, lines show the 10 most confident. Matching uses background-removed inputs, and the
matches are drawn on the original photos.}
\label{fig:match_examples}
\end{figure}

Figure~\ref{fig:match_examples} shows the correspondences of LoMa +
\our{} on correct top-1 retrievals, one per dataset. They link parts of
the coat, skin, or shell pattern across changes in viewpoint and
illumination, for example the spots of the hyena and the leopard, the
stripes of the nyala, and the yellow patches of the salamander. Because
the ranking is built from these correspondences, every retrieval can be
inspected. An expert can check which markings support a proposed match
before accepting it, which a similarity score from a global embedding
alone does not show. This matters in conservation monitoring, where a
wrong identity propagates into population estimates.

\FloatBarrier
\section{Conclusion}
\label{sec:conclusion}

We showed that general-purpose local feature matchers can be effectively specialized to wildlife re-identification using only the identity labels already available in standard monitoring datasets. Across eight datasets, adapting the matching module improves LoMa over its pretrained checkpoint and over a competitive local--global fusion method at almost every candidate budget, and experiments with RDD show that the approach extends beyond a single matcher architecture. The adapted matcher also uses identity supervision more effectively than closed-set classification and retains its gains on individuals never observed during training, indicating that it learns a transferable matching prior rather than memorizing identities. \our{} therefore provides a data-efficient way to specialize local correspondence models without keypoint or geometric annotations. Because the resulting model can be applied to newly observed individuals without retraining and exposes the correspondences underlying each retrieval, it is well suited to practical wildlife identification workflows. These results also suggest broader directions for combining domain-adapted local matching with learned global retrieval, and for training matchers that transfer across multiple species and monitoring domains.

\bibliographystyle{splncs04}
\bibliography{mybibliography}

\begin{thebibliography}{10}
\providecommand{\url}[1]{\texttt{#1}}
\providecommand{\urlprefix}{URL }
\providecommand{\doi}[1]{https://doi.org/#1}

\bibitem{adam2025overview}
Adam, L., Papafitsoros, K., Kovář, R., Čermák, V., Picek, L.: Overview of
  animalclef 2025: Recognizing individual animals in images. In: Working Notes
  of the Conference and Labs of the Evaluation Forum (CLEF 2025). CEUR Workshop
  Proceedings, Madrid, Spain (2025),
  \url{https://ceur-ws.org/Vol-4038/paper_231.pdf}

\bibitem{adam2025wildlifereid10k}
Adam, L., Čermák, V., Papafitsoros, K., Picek, L.: Wildlifereid-10k: Wildlife
  re-identification dataset with 10k individual animals. In: 2025 IEEE/CVF
  Conference on Computer Vision and Pattern Recognition Workshops (CVPRW). pp.
  2090--2100 (2025). \doi{10.1109/CVPRW67362.2025.00197}

\bibitem{bolger2012markrecapture}
Bolger, D., Morrison, T., Vance, B., Lee, D., Farid, H.: A computer-assisted
  system for photographic mark–recapture analysis. Methods in Ecology and
  Evolution  \textbf{3},  813--822 (10 2012).
  \doi{10.1111/j.2041-210X.2012.00212.x}

\bibitem{lilawphyena_2022}
{Botswana Predator Conservation Trust}: Hyena id 2022. LILA BC (May 2022),
  \url{https://lila.science/datasets/hyena-id-2022/}

\bibitem{lilawpleopard_2022}
{Botswana Predator Conservation Trust}: Leopard id 2022. LILA BC (May 2022),
  \url{https://lila.science/datasets/leopard-id-2022/}

\bibitem{carion2026sam3segmentconcepts}
Carion, N., Gustafson, L., Hu, Y.T., Debnath, S., Hu, R., Suris, D., Ryali, C.,
  Alwala, K.V., Khedr, H., Huang, A., Lei, J., Ma, T., Guo, B., Kalla, A.,
  Marks, M., Greer, J., Wang, M., Sun, P., Rädle, R., Afouras, T., Mavroudi,
  E., Xu, K., Wu, T.H., Zhou, Y., Momeni, L., Hazra, R., Ding, S., Vaze, S.,
  Porcher, F., Li, F., Li, S., Kamath, A., Cheng, H.K., Dollár, P., Ravi, N.,
  Saenko, K., Zhang, P., Feichtenhofer, C.: Sam 3: Segment anything with
  concepts (2026), \url{https://arxiv.org/abs/2511.16719}

\bibitem{chelak2021edendeepfeaturedistribution}
Chelak, I., Nepovinnykh, E., Eerola, T., Kälviäinen, H., Belykh, I.: Eden:
  Deep feature distribution pooling for saimaa ringed seals pattern matching
  (2021), \url{https://arxiv.org/abs/2105.13979}

\bibitem{Chen_2025_CVPR}
Chen, G., Fu, T., Chen, H., Teng, W., Xiao, H., Zhao, Y.: Rdd: Robust feature
  detector and descriptor using deformable transformer. In: Proceedings of the
  Computer Vision and Pattern Recognition Conference (CVPR). pp. 6394--6403
  (June 2025)

\bibitem{crall2013hotspotter}
Crall, J., Stewart, C., Berger-Wolf, T., Rubenstein, D., Sundaresan, S.:
  Hotspotter — patterned species instance recognition. In: Proceedings of
  IEEE Workshop on Applications of Computer Vision. pp. 230--237 (01 2013).
  \doi{10.1109/WACV.2013.6475023}

\bibitem{czarnota}
Czarnota, P., Loch, J., Armatys, P., Wierzbowska, I., Matysek, M.: The
  monitoring of eurasian lynx (lynx lynx) in the gorce national park
  (carpathians, poland) with the use of camera trapping between 2014-2018:
  recommendations on the future species protection. [monitoring rysia
  euroazjatyckiego (lynx lynx) z użyciem fotopułapek na terenie
  gorczańskiego parku narodowego - wskazania do ochrony gatunku w oparciu o
  wyniki z okresu 2014-2018]. Studia i Materiały CEPL w Rogowie
  \textbf{59}(2),  77--87 (2019)

\bibitem{dlamini2020automated}
Dlamini, N., Zyl, T.L.v.: Automated identification of individuals in wildlife
  population using siamese neural networks. In: 2020 7th International
  Conference on Soft Computing \& Machine Intelligence (ISCMI). pp. 224--228
  (2020). \doi{10.1109/ISCMI51676.2020.9311574}

\bibitem{dula}
Duľa, M., Bojda, M., Chabanne, D.B., Drengubiak, P., Ľuboslav Hrdý,
  Krojerová-Prokešová, J., Kubala, J., Labuda, J., Marčáková, L.,
  Oliveira, T., Smolko, P., Váňa, M., Kutal, M.: Multi-seasonal systematic
  camera-trapping reveals fluctuating densities and high turnover rates of
  carpathian lynx on the western edge of its native range. Scientific Reports
  \textbf{11}(9236),  1--12 (2021). \doi{10.1038/s41598-021-88348-8}

\bibitem{hoffer2018deepmetriclearningusing}
Hoffer, E., Ailon, N.: Deep metric learning using triplet network (2018),
  \url{https://arxiv.org/abs/1412.6622}

\bibitem{holmberg_norman_arzoumanian_2009}
Holmberg, J., Norman, B., Arzoumanian, Z.: Estimating population size,
  structure, and residency time for whale sharks rhincodon typus through
  collaborative photo-identification. Endangered Species Research  \textbf{7},
  39–53 (Apr 2009). \doi{10.3354/esr00186},
  \url{https://www.int-res.com/journals/esr/articles/esr00186}

\bibitem{kargin2026sparrta}
Kargin, T.C., Jasiński, W., Pardyl, A., Zieliński, B., Przewięźlikowski,
  M.: Sparrta: A synthetic benchmark for evaluating spatial intelligence in
  visual foundation models (2026), \url{https://arxiv.org/abs/2601.11729}

\bibitem{lindenberger2023lightgluelocalfeaturematching}
Lindenberger, P., Sarlin, P.E., Pollefeys, M.: Lightglue: Local feature
  matching at light speed (2023), \url{https://arxiv.org/abs/2306.13643}

\bibitem{lowe2004sift}
Lowe, D.G.: Distinctive image features from scale-invariant keypoints.
  International Journal of Computer Vision  \textbf{60},  91--110 (2004).
  \doi{10.1023/B:VISI.0000029664.99615.94}

\bibitem{nordstrom2026lomalocalfeaturematching}
Nordström, D., Edstedt, J., Bökman, G., Astermark, J., Heyden, A., Larsson,
  V., Wadenbäck, M., Felsberg, M., Kahl, F.: Loma: Local feature matching
  revisited (2026), \url{https://arxiv.org/abs/2604.04931}

\bibitem{otarashvili2024multispeciesanimalreidusing}
Otarashvili, L., Subramanian, T., Holmberg, J., Levenson, J.J., Stewart, C.V.:
  Multispecies animal re-id using a large community-curated dataset (2024),
  \url{https://arxiv.org/abs/2412.05602}

\bibitem{picek2025czechlynx}
Picek, L., Straka, J., Jirik, M., Belotti, E., Du{\v{l}}a, M., Krausov{\'a},
  J., Bojda, M., {\v{C}}erm{\'a}k, V., Bufka, L., Dvo{\v{r}}{\'a}k, R.,
  Hrd{\'y}, L., Kocourek, V., Labuda, J., Toman, L., Trul{\'\i}k, V.,
  V{\'a}{\v{n}}a, M., Kutal, M.: Czechlynx: A dataset for individual
  identification and pose estimation of the eurasian lynx. Scientific Data
  \textbf{13}(511),  1--11 (2026). \doi{10.1038/s41597-026-06853-9}

\bibitem{siméoni2025dinov3}
Siméoni, O., Vo, H.V., Seitzer, M., Baldassarre, F., Oquab, M., Jose, C.,
  Khalidov, V., Szafraniec, M., Yi, S., Ramamonjisoa, M., Massa, F., Haziza,
  D., Wehrstedt, L., Wang, J., Darcet, T., Moutakanni, T., Sentana, L.,
  Roberts, C., Vedaldi, A., Tolan, J., Brandt, J., Couprie, C., Mairal, J.,
  Jégou, H., Labatut, P., Bojanowski, P.: Dinov3 (2025),
  \url{https://arxiv.org/abs/2508.10104}

\bibitem{smith2026dsgtarcanimalclef2026}
Smith, E.S., Miyaguchi, A., Palamari, S., Evangelista, D.: Ds@gt arc at
  animalclef 2026: Species-aware graph construction for multi-species animal
  re-identification (2026), \url{https://arxiv.org/abs/2607.16453}

\bibitem{suessle}
Süßle, V., Heurich, M., Downs, C., Weinmann, A., Hergenröther, E.: Cnn based
  flank predictor for quadruped animal species. Workshop Camera Traps, AI and
  Ecology  (2024), \url{10.48550/arXiv.2406.13588}

\bibitem{wahltinez_wahltinez_2024}
Wahltinez, O., Wahltinez, S.J.: An open‐source general purpose machine
  learning framework for individual animal re‐identification using few‐shot
  learning. Methods in Ecology and Evolution  \textbf{15}(2),  373–387 (Jan
  2024). \doi{10.1111/2041-210x.14278},
  \url{https://besjournals.onlinelibrary.wiley.com/doi/10.1111/2041-210X.14278}

\bibitem{wang2026understandingoptimizingattentionbasedsparse}
Wang, Q.: Understanding and optimizing attention-based sparse matching for
  diverse local features (2026), \url{https://arxiv.org/abs/2602.08430}

\bibitem{zindi_2022}
Zindi: Turtle recall: Conservation challenge (2022),
  \url{https://zindi.world/competitions/turtle-recall-conservation-challenge}

\bibitem{wildfusion2025}
Čermák, V., Picek, L., Adam, L., Neumann, L., Matas, J.: Wildfusion:
  Individual animal identification with calibrated similarity fusion. In:
  Computer Vision -- ECCV 2024 Workshops. pp. 18--36. Springer Nature
  Switzerland (2025)

\bibitem{cermak2024wildlifedatasets}
Čermák, V., Picek, L., Adam, L., Papafitsoros, K.: Wildlifedatasets: An
  open-source toolkit for animal re-identification. In: 2024 IEEE/CVF Winter
  Conference on Applications of Computer Vision (WACV). pp. 5941--5951 (2024).
  \doi{10.1109/WACV57701.2024.00585}

\end{thebibliography}
\end{document}